\documentclass[9pt,twocolumn,letterpaper]{article}
\usepackage[
  top=0.6in,
  bottom=0.8in,
  left=0.65in,
  right=0.65in
]{geometry}
\usepackage{newtxtext,newtxmath}
\usepackage{graphicx}
\usepackage{subcaption}
\usepackage{booktabs}

\usepackage{amssymb}
\usepackage{xcolor}
\usepackage{multirow}

\usepackage{pifont}
\newcommand{\cmark}{\ding{51}} % checkmark
\newcommand{\xmark}{\ding{55}} % cross
\usepackage{url}
\usepackage{hyperref}
\usepackage{amsmath}
\usepackage[numbers]{natbib}
\begin{document}

\title{Automated In-the-Wild Data Collection for Continual AI Generated Image Detection}
\author{
Thanasis Pantsios, Dimitrios Karageorgiou, Christos Koutlis,\\
George Karantaidis, Olga Papadopoulou, Symeon Papadopoulos\\[0.5em]
Information Technology Institute, CERTH, Thessaloniki, Greece\\
\texttt{\{apantsios, dkarageo, ckoutlis, karantai, olgapapa, papadop\}@iti.gr}
}
\date{}
\maketitle
\begin{abstract}
The rapid advancement of generative Artificial Intelligence (AI) has introduced significant challenges for reliable AI-generated image detection. Existing detectors often suffer from performance degradation under distribution shifts and when encountering newly emerging generative models. In this work, we propose a data-centric continual adaptation framework for updating detectors in evolving environments. We show that both in-the-wild data and generator-driven data are essential for adapting detectors. We introduce an automated, weakly supervised pipeline for constructing in-the-wild datasets through fact-check article retrieval. Additionally, we demonstrate that incorporating even a small amount of generator-driven data during training enables effective adaptation to newly emerging models, while combining it with in-the-wild data within a continual learning framework enables robust adaptation and mitigates catastrophic forgetting. Extensive experiments on two state-of-the-art detectors show significant improvements of +9.14\% and +8\% in average accuracy, respectively. The proposed dataset and model checkpoints are publicly available at \url{https://mever-team.github.io/WildFC/}.

\end{abstract}

\section{Introduction}
\label{sec:intro}

Generative Artificial Intelligence (AI) is rapidly evolving and playing an important role across many sectors \cite{sengar2025generative}. Advances in generation quality have led to photorealistic images that humans often cannot reliably distinguish from real ones. Capabilities have also expanded beyond text-to-image generation to include editing, style transfer, and multimedia creation \cite{yazdani2025generative,chen2025opengpt}. User-friendly interfaces, open models and reduced costs have led to a massive increase in generated content, which increases the risk of %These technologies are used for beneficial applications, such as digital content creation, film production, healthcare, education, and accessibility, but also 
misuse for disinformation, fraud, non-consensual content, and political manipulation calling for reliable and content verification mechanisms \cite{sengar2025generative,yazdani2025generative}.

\begin{figure}[t]
    \centering
    \includegraphics[width=0.85\linewidth]{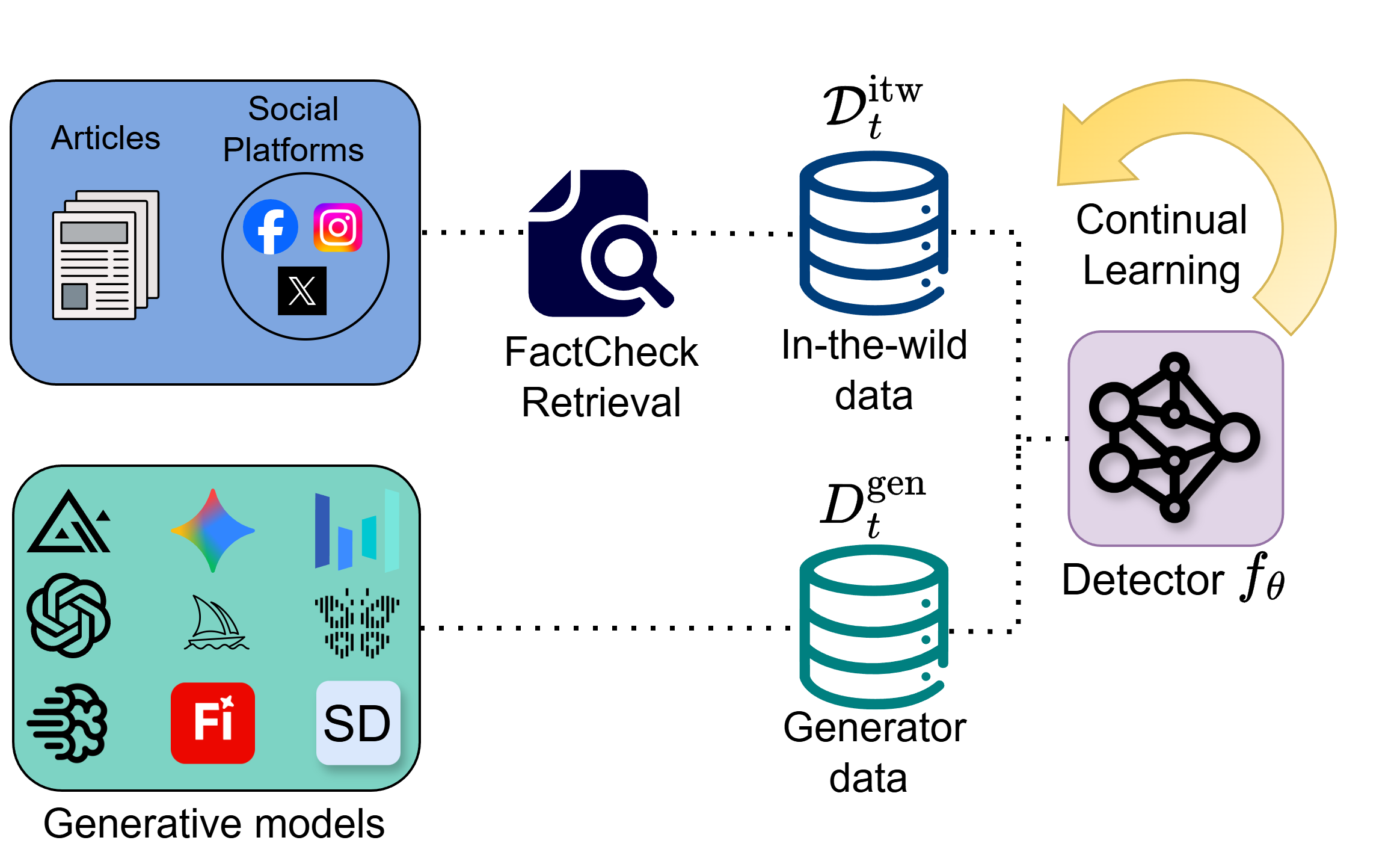}
    \caption{The proposed framework combines regular data collection from in-the-wild data $\mathcal{D}^{\text{itw}}_t$ via fact-check article retrieval and generator-driven data $\mathcal{D}^{\text{gen}}_t$ within a continual learning pipeline to adapt detectors $f_\theta$.}
    \label{fig:framework}
\end{figure}

In this context, AI-generated image detection (AID) has emerged as a critical challenge. Various approaches have been proposed, from early CNN-based detectors for GAN-generated images \cite{wang2020cnn} to recent approaches targeting diffusion models, including spatial-domain methods \cite{tan2024rethinking}, spectral analysis \cite{karageorgiou2025any}, and Vision-Language Models (VLMs) \cite{cozzolino2024raising,koutlis2024leveraging}. These perform well on controlled benchmarks, but struggle to generalize to out-of-distribution samples, particularly those from unseen generative models, post-processing operations, and real-world data \cite{yan2025sanity,konstantinidou2025navigating} leading to large performance drops in practical settings \cite{li2025bridging}.

As generative models evolve with new capabilities and user interactions, the distribution of AI-generated images is constantly changing, requiring continuous adaptation of detection models. Recent approaches address this challenge through continual and online learning frameworks \cite{wang2022s,doherty2024clofai,epstein2023online,azizpour2024e3,lu2025liteupdate}, primarily focusing on adapting detectors to new generators. However, many models are closed-source and computationally expensive, limiting access to representative training data. Moreover, controlled pipelines fail to capture post-processing, editing, and platform-specific transformations in real-world content.

To this end, we propose a continual data collection and learning framework for adapting detectors under distribution shift, as depicted in Fig.~\ref{fig:framework}. Our framework jointly leverages generator-driven and in-the-wild data, allowing detectors to adapt to both newly emerging models and real-world variations. While detector architectures improve, we argue that a data-centric perspective provides a necessary direction to address the evolving nature of AID. Our approach introduces a fact-check retrieval pipeline that enables automated, weakly supervised dataset construction by incorporating in-the-wild data from real-world sources. We show that even small amounts of such data significantly improve performance. We evaluate our framework on two state-of-the-art detectors, RINE \cite{koutlis2024leveraging} and SPAI \cite{karageorgiou2025any}, which remain vulnerable to distribution shifts and emerging generative models and represent diverse approaches, including spectral and CLIP-based methods, making them suitable for assessing robust generalization. In summary: %, our main contributions are:
\begin{itemize}
    \item We propose a continual data collection and learning framework that adapts detectors to evolving data distributions and generalizes across detector architectures.
    \item We introduce a fact-check retrieval pipeline for automated, weakly supervised creation of in-the-wild datasets for AID.
    \item We introduce an evolving dataset of AI-generated images, including 2,884 in-the-wild instances and 5,439 images generated by 19 recent generative models. In addition, we present a large dataset of 213,674 real images collected in the wild.
    \item We demonstrate the general applicability of our framework across two different detection models, consistently improving state-of-the-art performance by +9.14\% for SPAI and +8\% for RINE in average accuracy.
\end{itemize}

\section{Related Work}
\label{sec:related_work}

\begin{figure*}[t]
\centering
\includegraphics[width=0.7\textwidth]{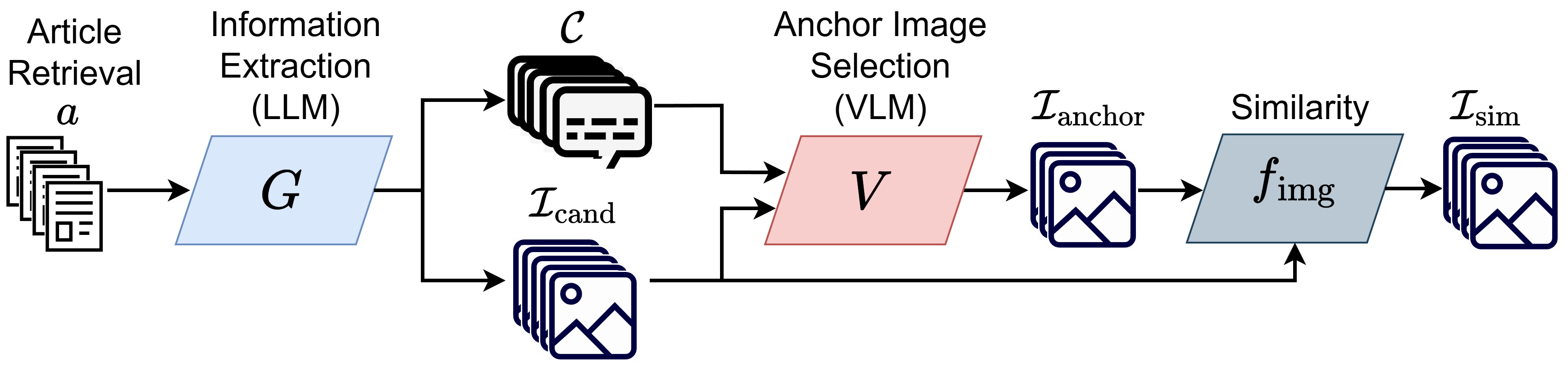}
\vspace{-3pt}
\caption{Fact-check retrieval pipeline: Given an article $a$, an LLM $\mathcal{G}$ extracts textual descriptions $\mathcal{C}$ referring to AI-generated images mentioned in $a$, along with a set of candidate images $\mathcal{I}_{\mathrm{cand}}$. A VLM $\mathcal{V}$ identifies anchor images $\mathcal{I}_{\mathrm{anchor}}$ that are semantically aligned with $\mathcal{C}$, and a similarity function based on an image encoder $f_{\mathrm{img}}$ further expands this set to $\mathcal{I}_{\mathrm{sim}}$.}
\label{fig:factcheck_diagram}
\end{figure*}

This section reviews advances in AID, including methods, datasets, and continual adaptation strategies. We highlight limitations in handling real-world data and evolving models, motivating continuous and adaptive  frameworks.

\subsection{AI-Generated Image Detection}

Early AID works focus primarily on GAN-based generators. Wang et al \cite{wang2020cnn} demonstrate that detectors trained on a limited set of generators can generalize to unseen ones, indicating shared generative artifacts.  Subsequent works explored more robust detection strategies. SAFE \cite{li2025improving} adopts a lightweight framework for both GAN- and diffusion-generated images, incorporating simple transformations and a local-awareness training strategy. CLIP-based approaches \cite{cozzolino2024raising} show strong generalization using penultimate-layer features and a simple SVM classifier. RINE \cite{koutlis2024leveraging} explores CLIP’s representation capabilities by leveraging intermediate Transformer blocks that capture low-level visual artifacts. It employs a linear mapping to project these features into a forgery-aware space and a trainable importance estimator to weight each block’s contribution. SPAI \cite{karageorgiou2025any} models the spectral distribution of real images using masked spectral learning and treats generated images as out-of-distribution samples through a spectral reconstruction module.

While detectors excel on controlled benchmarks, several works show significant degradation on realistic scenarios and unseen generators~\cite{karageogiou2024evolution}. Yan et al. \cite{yan2025sanity} posit that detectors often misclassify challenging AI-generated images, indicating that the problem is far from solved. Similarly, Konstantinidou et al. \cite{konstantinidou2025navigating} show that models trained on benchmark data struggle with real-world variations, and that incorporating in-the-wild data during training substantially improves performance. To address this challenge, B-Free \cite{guillaro2025bias} introduces a bias-free approach that constructs semantically aligned real–fake pairs via controlled generation, isolating synthesis artifacts and achieving strong generalization. While this improves data quality in a static setting, our work follows an orthogonal direction by addressing the temporal evolution of data distributions, modeling AID as a non-stationary problem and enabling continuous adaptation through in-the-wild and generator-driven data.

\subsection{Datasets and Benchmarks for AID}
Recent advances in the area highlight the importance of dataset design, including curated benchmarks, in-the-wild collections, and realistic evaluation protocols. Twigma \cite{chen2023twigma} introduces a real-world dataset  collected from Twitter, demonstrating their characteristics and temporal evolution. Corvi et al. \cite{corvi2023detection} introduce a dataset including diffusion-generated images to extend AID beyond GAN images, highlighting that different generative models produce distinct artifacts and showing performance degradation under realistic distortions. RealHD \cite{yu2025realhd} constructs a high-quality and diverse dataset for AID, addressing limitations of earlier benchmarks in image quality, prompt complexity and diversity. RealHD does not rely solely on text-to-image pipelines but includes multiple generation settings such as inpainting, refinement, and face swapping, with images generated by state-of-the-art diffusion models. Building on these directions, we propose a unified update framework that integrates both in-the-wild and generator-driven data, moving beyond static datasets toward a continuously evolving data distribution.

While not specifically designed for AID, recent works explore structured and synthetic data generation for well-curated datasets. OpenGPT-4o-Image \cite{chen2025opengpt} introduces a comprehensive dataset with a hierarchical task taxonomy and automated data generation using GPT-4o, emphasizing structured and diverse data construction. Similarly, Echo-4o-Image \cite{ye2025echo} is a synthetic dataset generated by GPT-4o, providing clean and controllable supervision while covering rare and long-tail scenarios. In addition, RealBench \cite{ye2025realgen} is a benchmark for evaluating the photorealism of generated images. The work introduces a photorealistic text-to-image framework that employs an adversarial approach through a detector reward mechanism, highlighting the evolving nature of the generation process. These directions are important for advancing dataset design for AID, improving data construction and curation procedures.

A growing line of work focuses on in-the-wild benchmarks to reflect realistic AID conditions, highlighting the gap between controlled benchmarks and practical deployment \cite{yan2025sanity, papadopoulou2025synthetic, li2025artificial, konstantinidou2025navigating}. AI-GenBench \cite{pellegrini2025ai} introduces a temporal benchmark with a chronologically organized dataset, combining multiple existing real and synthetic datasets. In this setting, detectors are trained on earlier generative models and evaluated on newer ones, emphasizing the evolving nature of the problem. RRDataset \cite{li2025bridging} is introduced as a benchmark designed to evaluate AID under realistic conditions, incorporating diverse scenarios, social media transmission effects, and re-digitization processes, and revealing significant limitations of existing detectors. Our work addresses these challenges by incorporating real-world data through a continuously evolving dataset, while enabling adaptation to emerging generative models through a continual update framework for detectors.

%AIGI-Bench \cite{li2025artificial}

\subsection{Continual and Online Adaptation in AID}

Several works explore AID under continual learning. S-Prompts \cite{wang2022s} propose a prompt-based method where task-specific prompts are learned for sequential domains while keeping the backbone model frozen. CLOFAI \cite{doherty2024clofai} introduces a domain-incremental AID benchmark where data are organized into sequential tasks reflecting the evolution of generative models. However, both approaches assume discrete task boundaries, where each incremental step is tied to a specific generative model, which does not reflect the evolving nature of real-world scenarios.

Recent works explore adaptive AID frameworks. Epstein et al. \cite{epstein2023online} propose an online detection framework, providing insights into temporal generalization through a streaming setup where detectors are trained on a set of generators and evaluated on future unseen ones. Similarly, E3 \cite{azizpour2024e3} presents an adaptive approach that addresses newly emerging generators with limited data by learning generator-specific expert embedders and combining them through a fusion mechanism. LiteUpdate \cite{lu2025liteupdate} introduces a lightweight strategy that updates detectors using small task-specific modules, along with a representative sample selection mechanism. It also incorporates a model merging scheme to balance adaptation to new generators while mitigating catastrophic forgetting. Pellegrini et al. \cite{pellegrini2025generalized} offer a comprehensive analysis of architecture-agnostic design choices for  detectors, including augmentation, preprocessing, multiclass supervision, and incremental learning. For continual adaptation, they explore replay-based methods such as harmonic and class-balanced replay across successive generator windows. %, demonstrating that replay can effectively balance adaptation and retention. 
However, these approaches remain generator-centric and rely on updates tied to individual generative models, emphasizing architectural adaptation without explicitly incorporating in-the-wild data.
\section{Methodology}
\label{sec:method}

\subsection{Problem Formulation}

AID is a binary classification problem defined over an input space $\mathcal{X}$ (images) and label space $\mathcal{Y} = \{0,1\}$, where $y=1$ denotes AI-generated images and $y=0$ real images. The data is generated according to a joint distribution $P_t(x,y)$ that evolves over time $t$. We consider a non-stationary environment, where the underlying data distribution changes sequentially over time:
\begin{equation}
P_{t_0}(x,y) \neq P_t(x,y), \quad \text{for } t > t_0.
\end{equation}

This temporal evolution violates the i.i.d. assumption and leads to dataset shift \cite{moreno2012unifying}, making AID a rapidly evolving problem. The first source of distribution shift arises from the continuous development of new generative models, architectures, and synthesis pipelines. These introduce new patterns and artifacts, modifying their distribution over time. In addition, the distribution shift arises from changes in user behavior and real-world usage over time. As generative models are adopted in practice, users interact with them in diverse ways (e.g., editing, compositional generation, platform-specific processing), which affects how synthetic content is produced and perceived. Formally, this corresponds to covariate shift:

\begin{equation}
P_t(x \mid y) \neq P_{t_0}(x \mid y), \quad
P_t(y \mid x) = P_{t_0}(y \mid x).
\end{equation}

These two complementary mechanisms motivate the need for multiple data sources to track the evolving distribution. In particular, our framework leverages generator-driven data to capture covariate shift induced by evolving generative models, and in-the-wild data to capture covariate shift from real-world usage dynamics. This dual-source strategy enables a better approximation of the distribution $P_t(x,y)$ and supports effective updating of detectors.

\subsection{Fact-Check Retrieval}
\label{factcheck_retrieval_method}
Our framework constructs a dataset of AI-generated images by leveraging fact-check articles, as depicted in Fig.~\ref{fig:factcheck_diagram}. Fact-check articles are treated as sources of weak supervision describing AI-generated images. 

\begin{figure}[b]
\centering
\includegraphics[width=0.75\linewidth]{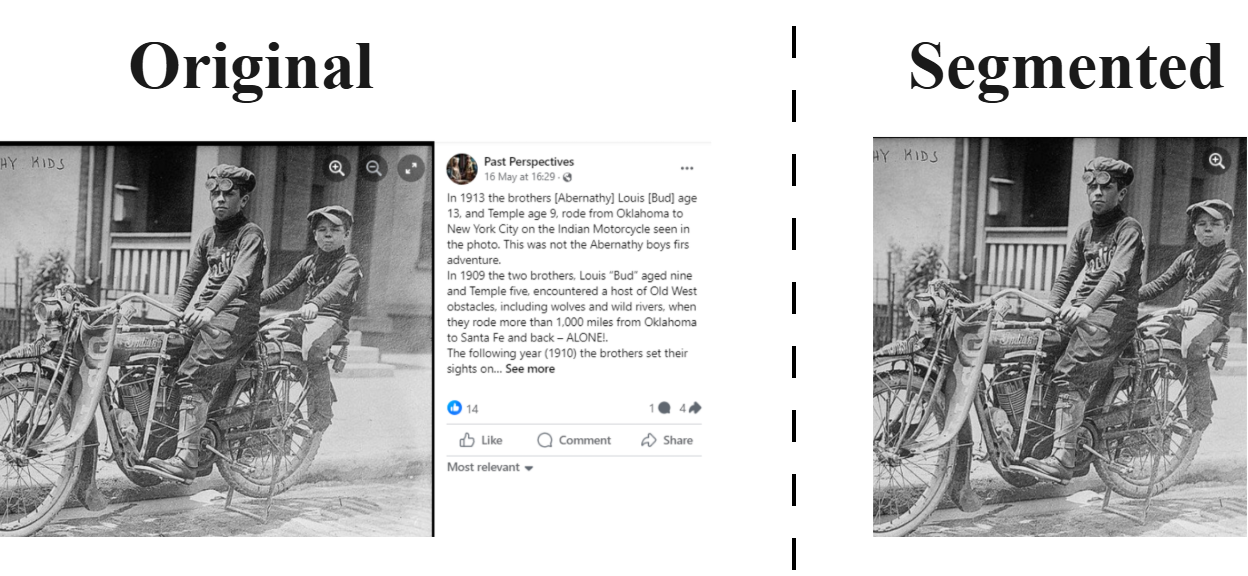}
\caption{Image segmentation: original (left) and segmented (right) image.}
\label{fig:ex_seg}
\end{figure}

\begin{figure}[b] 
\centering 
\includegraphics[width=0.8\linewidth]{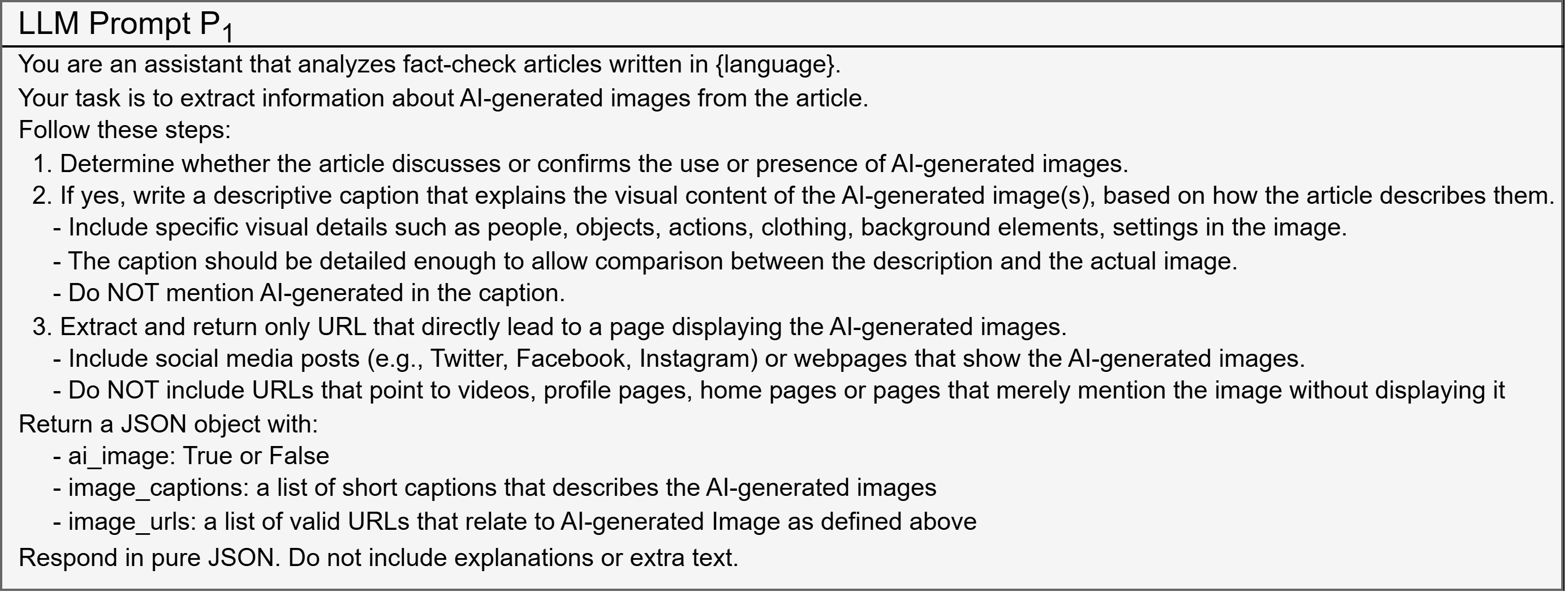} 
\vspace{-3pt}
\caption{Instruction prompt $p_1$ template for LLM $\mathcal{G}$.}
\label{fig:prompt_1} 
\vspace{-3pt}
\end{figure}

An information extraction process is performed using a large language model $\mathcal{G}$, which analyzes an article $a$ and extracts textual descriptions corresponding to AI-generated images mentioned in the article narrative. Articles that do not refer to AI-generated images are discarded. Formally, this process is defined as

\begin{equation}
\mathcal{C} = \mathcal{G}(a, p_1) = \{c_i\}_{i=1}^{K}
\end{equation}
where  $p_1$ is a task-specific instruction prompt, as depicted in Fig.~\ref{fig:prompt_1}, and $\mathcal{G}$ represents the LLM-based extraction function. 

The set $\mathcal{C}$ contains the textual descriptions derived from article $a$. Each description $c_i$ corresponds to the visual content of the $i$-th AI-generated image referenced in the article, while $K$ denotes the number of extracted descriptions.

For each article $a$, we collect a set of candidate images $\mathcal{I_{\mathrm{cand}}} = \{x_j\}_{j=1}^{N}$ associated with the article, where $x_j$ denotes the $j$-th candidate image and $N$ represents the number of collected images. The goal of this process is to automatically determine which of the candidate images correspond to the AI-generated content described in the article and which are unrelated and should be discarded.

To this end, an anchor image selection process is performed to identify images that correspond to the extracted textual descriptions. To achieve this, we employ a VLM that evaluates the semantic compatibility between candidate images and textual descriptions.

\begin{figure}[t] 
\centering 
\includegraphics[width=0.8\linewidth]{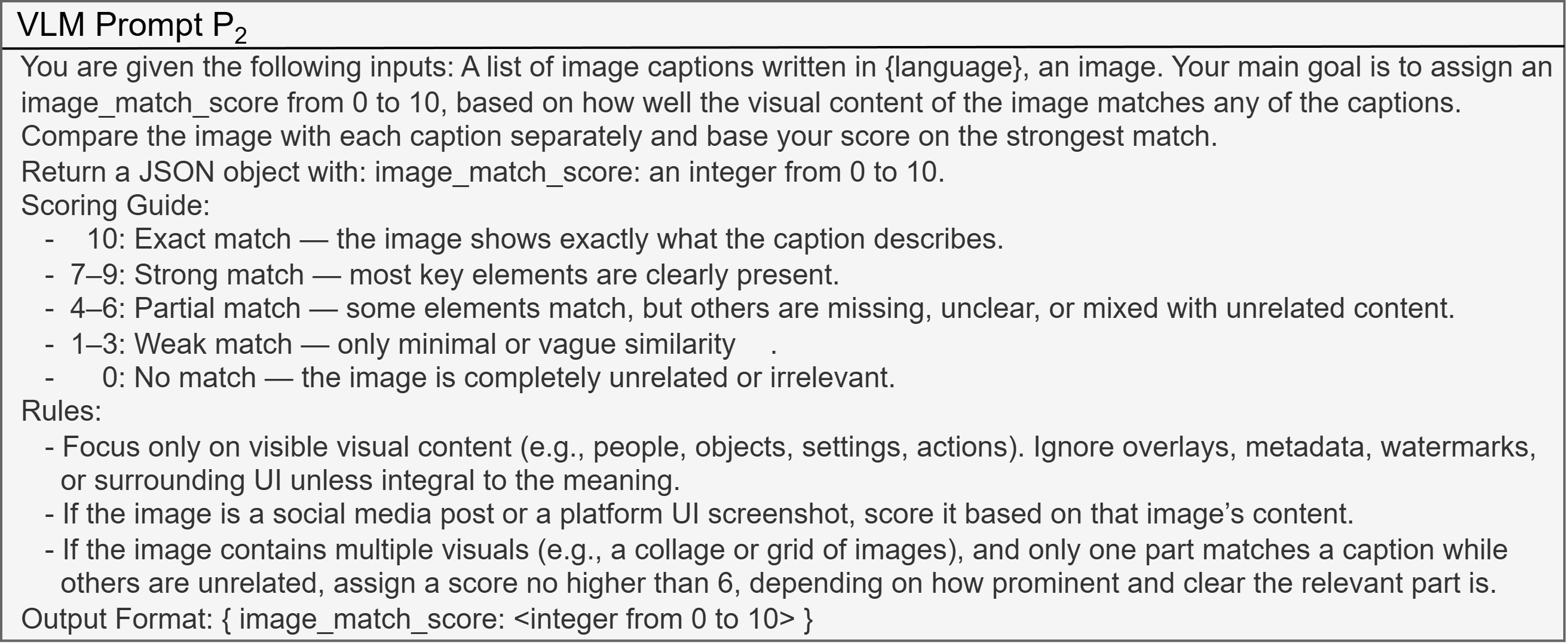} 
\vspace{-3pt}
\caption{Instruction prompt $p_2$ template for VLM $\mathcal{V}$.} 
\label{fig:prompt_2} 
\vspace{-3pt}
\end{figure}

Given a candidate image $x_j$ and a textual description $c_i$, the VLM produces a similarity score measuring the alignment between the image and the description. Since an article may describe multiple AI-generated images, the final similarity score for candidate image $x_j$ is defined as the maximum similarity across all textual descriptions:
\begin{equation}
s_{j} = \max_{c_i \in \mathcal{C}(a)} \mathcal{V}(x_j, c_i, p_2)
\end{equation}
where $\mathcal{V}$ denotes the VLM scoring function and $p_2$ is the task-specific prompt, as detailed in Fig.~\ref{fig:prompt_2}.

Candidate images whose similarity scores exceed a predefined anchor selection threshold $\tau_{\text{anchor}}$ are considered anchor images:

\begin{equation}
\mathcal{I}_{\mathrm{anchor}}
=
\{x_j \in \mathcal{I}_{\mathrm{cand}} \mid s_j \ge \tau_{\text{anchor}}\}
\end{equation}
To further expand the anchor set, we compute the visual similarity between anchor and candidate images using an image encoder $f_{\mathrm{img}}$:

\begin{equation}
s(x_a, x_j) =
\mathrm{sim}\big(
f_{\mathrm{img}}(x_a),
f_{\mathrm{img}}(x_j)
\big),
\end{equation}
where $x_a \in \mathcal{I}_{\mathrm{anchor}}$ and $x_j \in \mathcal{I}_{\mathrm{cand}}$.

\begin{figure}[t]
\centering
\includegraphics[width=0.6\linewidth]{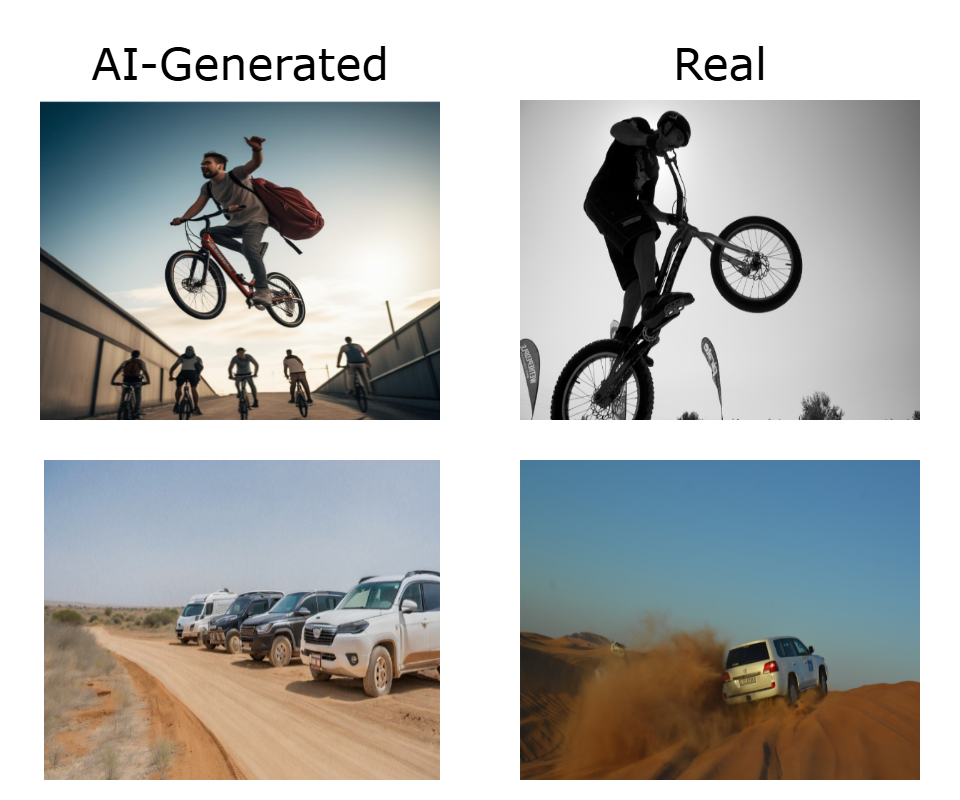}
\vspace{-3pt}
\caption{Examples of semantically aligned real–fake pairs.}
\label{fig:fake_real_pairs}
\vspace{-3pt}

\end{figure}

Candidate images are selected if their similarity with at least one anchor image exceeds a predefined threshold $\tau_{\text{sim}}$:

\begin{equation}
\mathcal{I}_{\mathrm{sim}}
=
\{x_j \in \mathcal{I_{\mathrm{cand}}} \mid
\max_{x_a \in \mathcal{I}_{\mathrm{anchor}}} s(x_a, x_j) \ge \tau_{\text{sim}}
\}
\end{equation}

The final image set associated with article $a$ is defined as

\begin{equation}
\mathcal{I}_{\mathrm{final}}
=
\mathcal{I}_{\mathrm{anchor}}
\cup
\mathcal{I}_{\mathrm{sim}}
\end{equation}

Additionally, to isolate candidate visual regions corresponding to the AI-generated content from complex images (e.g., social media screenshots), we apply an image segmentation process, producing segmented regions. An example is shown in Fig.~\ref{fig:ex_seg}. The segmented dataset is therefore defined as

\begin{equation}
\mathcal{I}_{\mathrm{seg}}
=
\bigcup_{x \in \mathcal{I}_{\mathrm{final}}} \mathcal{S}(x)
\end{equation}
where $\mathcal{S}(x)$ denotes the set of visual segments.

Finally, to construct semantically aligned real--fake image pairs, we retrieve real images from a pool of real images $\mathcal{R}$ that are visually similar to the AI-generated images in the final set. Given an AI-generated image $x_f \in \mathcal{I}_{\mathrm{seg}}$ and a real image $r_k \in \mathcal{R}$, we compute image similarity using the same image encoder:
\begin{equation}
s(x_f, r_k) =
\mathrm{sim}\big(
f_{\mathrm{img}}(x_f),
f_{\mathrm{img}}(r_k)
\big)
\end{equation}

The TopK retrieval is performed without replacement to ensure that each real image is selected at most once for a given AI-generated image. The final dataset consists of AI-generated images paired with semantically similar real images:
\begin{equation}
\mathcal{R}_{\mathrm{match}}(x_f)
=
\operatorname{TopK}_{r_k \in \mathcal{R}} s(x_f, r_k)
\end{equation}

\subsection{Continual Adaptive Framework}
We propose a general continual data collection and learning framework for adapting detectors under distribution shift, as illustrated in Fig.~\ref{fig:framework}. At regular update intervals, the framework incorporates three sources: newly collected in-the-wild data, samples from recently released generative models, and a continual learning mechanism.

At each update round $t$, we collect two complementary datasets. First, we gather in-the-wild data $\mathcal{D}^{\text{itw}}_t$ (e.g., WildFC; see Sec.~\ref{sec:fact_check_collection}), which reflects real-world image distributions and evolving artifact patterns. Second, we construct a dataset of recent generators $\mathcal{D}^{\text{gen}}_t$ (e.g., AIGenImages2026; see Sec.~\ref{sec:AIGenImages2026}). To mitigate catastrophic forgetting, we maintain a replay buffer $\mathcal{M}_{t-1}$, constructed by sampling a fixed proportion $\rho$ of the accumulated data observed up to round $t$. The training set at round $t$ is defined as:
\begin{equation}
\mathcal{D}_t = \mathcal{D}^{\text{itw}}_t \cup \mathcal{D}^{\text{gen}}_t \cup \mathcal{M}_{t-1}.
\label{eq:update_data}
\end{equation}

Given the aggregated dataset $\mathcal{D}_t$, the detector is updated by minimizing a loss function specific to the detector architecture:
\begin{equation}
\theta_t = \arg\min_{\theta} \; \mathcal{L}(f_\theta, \mathcal{D}_t).
\label{eq:update}
\end{equation}

This framework enables effective continuous adaptation. In-the-wild data enhances robustness to real-world distribution shifts, while generator-specific data facilitates rapid adaptation to emerging models. A continual learning component mitigates catastrophic forgetting, ensuring stable performance. While we instantiate this framework with a specific data collection pipeline and a replay-based strategy, its modular design remains compatible with alternative collection and continual learning methods.

\begin{table}[b]
\centering
\small
\setlength{\tabcolsep}{3pt}
\resizebox{0.85\columnwidth}{!}{
\begin{tabular}{lcccc}
\toprule
\textbf{Model} & \textbf{Date} & \textbf{Size} & \textbf{Split (Train/Test)} \\
SDXL & 2023-11 & 305 & 274 / 31 \\
FLUX.1 [pro] v1.1, SD 3.5 Med & 2024-10 & 305 & 274 / 31 \\
Reve Img 1.0, HiDream I1 Dev & 2025-03 & 305 & 274 / 31 \\
GPT-Img 1, Ideogram 3  & 2025-04 & 305 & 274 / 31 \\
Midjourney v7 & 2025-04 & 301 & 270 / 31 \\
Imagen 4 & 2025-05 & 305 & 274 / 31 \\
Gemini 2.5 Flash Img & 2025-10 & 305 & 274 / 31 \\
Firefly Img 5 & 2025-10 & 150 & 133 / 17 \\
FLUX.2, Z Img Turbo & 2025-11 & 305 & 275 / 31 \\
FLUX.2 [pro] & 2025-11 & 305 & 274 / 31 \\
FLUX.2 [dev] & 2025-11 & 205 & 182 / 23 \\
Gemini 3 Pro Img & 2025-11 & 305 & 276 / 31 \\
FLUX.2 [max] & 2025-12 & 205 & 182 / 23 \\
GPT-Img 1.5, Seedream 4.5 & 2025-12 & 305 & 274 / 31 \\
\bottomrule
\end{tabular}}
\caption{Overview of the AIGenImages2026 dataset.}
\label{tab:generators}
\end{table}

\begin{table*}[t]
\centering
\scriptsize
\setlength{\tabcolsep}{3.0pt}
\caption{Comparison of all detectors across benchmark datasets. Each entry reports AUC / ACC (\%). For each dataset and each metric, the best result is highlighted in bold, while the second best is underlined.}
\vspace{-5pt}
\label{tab:full_comparison}
\resizebox{0.9\textwidth}{!}{
\begin{tabular}{lcccccccccc}
\toprule
\textbf{Method}
& \multicolumn{5}{c}{\textbf{Recent Generators}}
& \multicolumn{4}{c}{\textbf{ITW Data}}
& \textbf{} \\
\cmidrule(lr){2-6} \cmidrule(lr){7-10}

& \textbf{AIGenImages2026}
& \textbf{MNW}
& \textbf{Echo-4o}
& \textbf{Nano-Consistent}
& \textbf{Synthbuster}
& \textbf{Chameleon}
& \textbf{DeepFake-Eval}
& \textbf{ITW-SM}
& \textbf{MediaEval}
& \textbf{AVG} \\
\midrule

CNN Detect \cite{wang2020cnn}
& 45.59 / 49.91
& 50.60 / 49.91
& 53.12 / 52.60
& 40.77 / 48.85
& 38.30 / 49.09
& 48.40 / 56.99
& 48.71 / 39.03
& 47.90 / 49.99
& 49.54 / 49.89
& 46.99 / 49.58 \\

NPR \cite{tan2024rethinking}
& 68.28 / 59.84
& 66.31 / 60.29
& 60.43 / 56.60
& 48.48 / 47.74
& 47.32 / 53.51
& 65.35 / 61.33
& 55.80 / 57.37
& 56.23 / 54.85
& 63.59 / 59.34
& 59.09 / 56.76 \\

SAFE \cite{li2025improving}
& 71.11 / 69.68
& 47.86 / 49.52
& 79.44 / 65.60
& 45.22 / 48.95
& 59.60 / 52.73
& 57.16 / 59.14
& 52.40 / 43.36
& 49.53 / 49.86
& 53.40 / 50.24
& 57.30 / 54.34 \\

LaDeDa \cite{cavia2024real}
& 60.62 / 56.08
& 59.39 / 56.55
& 79.24 / 72.55
& 44.58 / 45.30
& 62.93 / 59.99
& 65.11 / 61.86
& 62.54 / 52.11
& 74.48 / 67.21
& 69.91 / 63.76
& 64.31 / 59.49 \\

RINE \cite{koutlis2024leveraging}
& 75.70 / 60.82
& 84.41 / 66.98
& 75.70 / 68.50
& \textbf{85.58} / 60.07
& 83.76 / 83.01
& 39.27 / 44.77
& 71.84 / 49.45
& 70.21 / 56.18
& 64.71 / 54.65
& 72.35 / 60.49 \\

ITW\_RINE \cite{konstantinidou2025navigating}
& 90.80 / 71.74
& 93.50 / 81.88
& 50.89 / 48.75
& 68.81 / 63.81
& 90.87 / 73.42
& 88.19 / \underline{79.60}
& 84.56 / 75.56
& 96.23 / 81.49
& 89.08 / 76.41
& 83.66 / 72.52 \\

ITW\_SPAI \cite{konstantinidou2025navigating}
& 90.05 / 64.67
& 92.00 / 75.92
& \textbf{86.40} / 60.40
& 74.81 / 55.80
& \textbf{97.45} / \textbf{91.94}
& \underline{90.14} / 79.09
& \underline{88.73} / 71.50
& \textbf{98.13} / \textbf{92.81}
& \underline{94.79} / \underline{86.91}
& \underline{90.28} / 75.45 \\

RINE (ours)
& \textbf{97.75} / \textbf{92.13}
& \textbf{95.74} / \textbf{88.59}
& \textbf{88.45} / \underline{76.80}
& 78.35 / \underline{71.01}
& 94.81 / 77.75
& 89.64 / 78.58
& 87.59 / \underline{78.63}
& 95.74 / 82.50
& 91.09 / 78.80
& 91.02 / \underline{80.53} \\

SPAI (ours)
& \underline{95.93} / \underline{89.09}
& \underline{93.90} / \underline{84.08}
& 86.27 / \textbf{79.75}
& \underline{84.31} / \textbf{71.43}
& \underline{95.54} / \underline{89.86}
& \textbf{92.38} / \textbf{84.90}
& \textbf{90.09} / \textbf{82.33}
& \underline{97.72} / \underline{90.65}
& \textbf{94.83} / \textbf{89.20}
& \textbf{92.33} / \textbf{84.59} \\

\bottomrule
\end{tabular}}
\end{table*}

\section{Data collection}
\label{sec:data_collection}

\subsection{Fact-Checked Collection Retrieval}
\label{sec:fact_check_collection}

We introduce WildFC, an evolving dataset of in-the-wild AI-generated images, currently comprising 2,884 images collected in 2025 through the automated fact-check retrieval pipeline described in Sec.~\ref{factcheck_retrieval_method}. In total, 3,841 articles were retrieved from fact-checking databases and resources, including the Google Fact Check Tools \cite{google_factcheck_api} and the Database of Known Fakes (DBKF) \cite{dbkf}, using queries related to AI-generated content. Among these, 1,539 articles were identified as relevant using the Qwen3-8B-FP8 \cite{qwen3technicalreport} language model, which was used to detect references to AI-generated images, extract image URLs, and generate corresponding captions.

From the selected articles, a total of 10,387 candidate images were collected using web scraping tools such as Crawl4AI \cite{crawl4ai} and gallery-dl \cite{gallerydl}. For AI-generated image selection, we employed the Qwen2.5-VL-7B-Instruct \cite{qwen25_vl} VLM, which filtered the dataset to 2,884 relevant images. The set consisted mainly of JPEG files at 73.86\%, followed by PNG at 22.61\%, and WEBP at 3.54\%, spanning a resolution range of 0.06 MP to 16.78 MP. Finally, to isolate meaningful visual regions and remove platform-specific overlays or UI elements, we applied image segmentation using Grounding DINO Tiny \cite{liu2024grounding}. This process produced 2,298 segmented image samples, which serve as augmented views of the original content. 

A validation set of $300$ images, representing $10.40\%$ of the dataset, was examined to assess this weakly supervised automated labeling. The images were manually annotated, and a precision of $91.95\%$ was achieved. Although a small amount of noise was observed, it mainly occurs in more complex cases, such as ambiguous instances where the fact-checker could not provide a definitive verdict, or real images reused out of context to depict new events. The results presented in Sec.~\ref{sec:experiments} show that this noise is limited and that our dataset improves the performance of detectors.

\subsection{ Data from Recent Generators}
\label{sec:AIGenImages2026}
We construct AIGenImages2026, a dataset of images generated by 19 recent text-to-image models, as depicted in Table~\ref{tab:generators}. As our framework targets lightweight and cost-effective adaptation, we limit the number of samples to a maximum of 305 images per generator. We reserve 10\% of the generated data exclusively for evaluation.
Following recent trends in synthetic data generation \cite{chen2025opengpt,ye2025echo,ye2025realgen}, the prompts are designed to ensure both diversity and realism. We employ three groups of prompts with different perspectives and scopes. First, we generate 105 realistic text-to-image prompts using GPT-5.2 \cite{gpt52}, covering 21 real-world categories. Second, we include 100 prompts targeting more complex generation settings, including spatial reasoning, compositional constraints, and stylistic variations based on the OpenGPT-4o \cite{chen2025opengpt} dataset. Third, we incorporate 100 prompts derived from the VisualNews \cite{liu2021visual} dataset by selecting captions from representative topics and refining them using GPT-5.2, where identities and sensitive attributes are neutralized and visual semantics preserved. In total, the dataset contains 5,439 generated images, with 4,880 images used for training and 559 for testing. Most images are generated through the fal.ai API \cite{falai}, while a small subset is collected manually for models requiring separate access. For certain models, such as Midjourney v7, FLUX.2 [dev], and Adobe Firefly Image 5, fewer samples are available due to access limitations and API restrictions, while all models remain consistently represented across training and evaluation.

\subsection{Real Data Collection}
Real in-the-wild data is crucial for effectively updating detection models. We collect data from two primary sources. First, real images are gathered from news websites using the NewsAPI \cite{newsapi}, covering over 100 official media outlets across multiple categories, including general, science, health, entertainment, technology, business, and sports. Articles related to AI-generated content are explicitly filtered out using query parameters. Second, additional images are collected from social media platforms such as Facebook, Instagram, and X by monitoring 25 official news media accounts. The diversity of platforms and sources is essential for constructing a non-biased training set that reflects real-world data distributions.

In total, we construct a dataset of 213,674 real images collected between August and December 2025, with 155,761 images sourced from news outlets and 57,913 from social media. This continuously evolving dataset serves as a candidate pool from which semantically similar real images are retrieved for each AI-generated sample using a similarity-based matching strategy, forming aligned real–fake training pairs, as depicted in Fig.~\ref{fig:fake_real_pairs}. This process results in a balanced, challenging, and semantically consistent training dataset.

\section{Experiments}

\label{sec:experiments}

\subsection{Evaluation Datasets and Metrics}

We evaluate our approach on datasets covering both recent generators and in-the-wild conditions. For recent generators, we use a held-out test set of 558 images from AIGenImages2026 (see Table~\ref{tab:generators}, MNW \cite{MNW2025}, an evolving benchmark covering 51 generators, and Synthbuster \cite{bammey2023synthbuster}. Additionally, we include a publicly released subset of 5,000 images from Nano-Consistent \cite{ye2025echo} and 1,000 images from Echo-4o \cite{ye2025echo} generated with the GPT-4o model. For in-the-wild data, we use the challenging Chameleon \cite{yan2025sanity}, and ITW-SM \cite{konstantinidou2025navigating}, an in-the-wild dataset of 10,000 real and synthetic images collected from four social media platforms. Additionally, we use MediaEval-ITW \cite{papadopoulou2025synthetic}, which combines real images from datasets with synthetic images augmented through realistic transformations and 1,975 images from Deepfake-Eval \cite{chandra2025deepfake}, a multimodal benchmark based on real-world content. This combination enables evaluation across both controlled generation and realistic deployment scenarios.

We evaluate detection performance using two standard metrics for AID, Accuracy (ACC) and the area under the receiver operating characteristic curve (AUC). AUC measures the overall ranking capability independent of a fixed decision threshold, providing a global view of its discriminative ability. ACC reflects performance at a specific operating point and therefore provides a more practical measure. ACC is calculated using a threshold of 0.5.

\subsection{Implementation details}
For fact-check retrieval, we use $\tau_{\text{anchor}}=0.8$, $\tau_{\text{sim}}=0.75$, TopK $=500$, and a segmentation threshold of $0.4$, with CLIP ViT-L/14 \cite{radford2021learning} as the image encoder $f_{\mathrm{img}}$. Compared methods are evaluated using checkpoints from the official repositories, using the SIDBench \cite{schinas2024sidbench} framework for CNNDetect, NPR, and RINE, and AIGI-Bench \cite{li2025artificial} for SAFE and LaDeDa trained on WildRF \cite{cavia2024real}. Additionally, we evaluate the versions of RINE and SPAI proposed in \cite{konstantinidou2025navigating}, adopting the best-performing reported configuration with a DINOv2 \cite{oquab2023dinov2} backbone trained on TWIGMA \cite{chen2023twigma} and LDM \cite{corvi2023detection}. 

Experiments are conducted on the RINE and SPAI models proposed in \cite{konstantinidou2025navigating}, trained on WildFC, AIGenImages2026, and real images (see Sec. \ref{sec:data_collection}), while employing a replay memory $\rho$ of $5\%$. RINE is trained for $1$ epoch with a learning rate of $1 \times 10^{-3}$ and a batch size of $16$ on an NVIDIA RTX $4090$ GPU. SPAI is trained for $3$ epochs with a learning rate of $2.5 \times 10^{-7}$, using a batch size of $24$, mixed-precision training, and gradient accumulation over $2$ steps implemented in PyTorch \cite{paszke2019pytorch} on an NVIDIA RTX $5090$ GPU.

\subsection{Main Analysis}

A set of state-of-the-art detectors is evaluated to analyze the impact of distribution shift across recent generator data and in-the-wild data, as summarized in Table~\ref{tab:full_comparison}. Standard detectors such as CNN Detect \cite{wang2020cnn}, NPR \cite{tan2024rethinking}, and SAFE \cite{li2025improving} exhibit limited generalization ability, achieving average accuracies of $49.58\%$, $56.76\%$, and $54.34\%$, respectively. While LaDeDa \cite{cavia2024real} and RINE reach higher averages of $59.49\%$ and $60.49\%$, performance remains inconsistent across domains. Notably, RINE degrades on in-the-wild datasets, falling to $44.77\%$ on Chameleon and $49.45\%$ on Deepfake-Eval. These results indicate that models trained on static data distributions fail to generalize under evolving conditions.

Methods trained on in-the-wild data, such as ITW\_RINE and ITW\_SPAI, show improved robustness on real-world data, achieving $81.49\%$ and $92.81\%$ accuracy on ITW-SM, respectively. However, they exhibit degraded performance on Echo-4o and Nano-Consistent with $48.75\%$ and $63.81\%$ accuracy, highlighting the imbalance between robustness to real-world data and generalization to newly emerging generators.

Our framework yields consistent improvements across both domains. For RINE, average accuracy increases from $72.52\%$ to $80.53\%$, an absolute gain of $8.01\%$. Performance on recent generators improves significantly, where Echo-4o accuracy rises from $48.75\%$ to $76.80\%$ and Nano-Consistent from $63.81\%$ to $71.01\%$, while maintaining strong performance on in-the-wild datasets. Similar behavior is observed for SPAI, which achieves a $9.14\%$ gain, reaching an average accuracy of $84.59\%$, with improvements across both recent generators and in-the-wild datasets. These results demonstrate that static training is insufficient under evolving data distributions and highlight the importance of a continual, data-centric training pipeline to maintain robust performance over time. Our framework enables balanced adaptation, improving generalization to recent generators while preserving robustness on in-the-wild data.

\begin{figure}[t]
    \centering
    \includegraphics[width=0.8\linewidth]{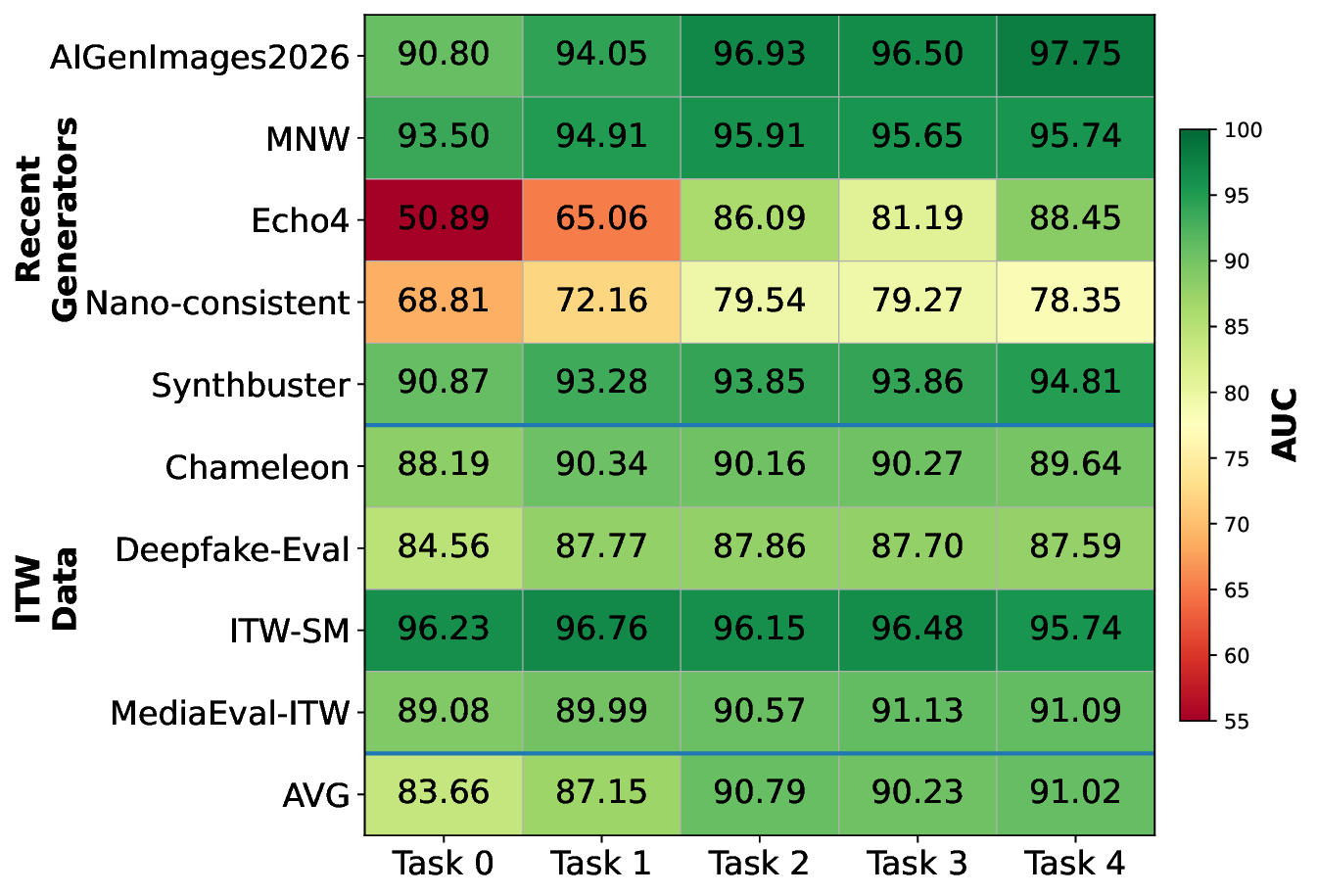}
    \vspace{-5pt}
    \caption{AUC performance under continual learning. Results
demonstrate adaptation to generators while maintaining and
improving performance on in-the-wild data across tasks.}
    \label{fig:continual_auc}
    \vspace{-10pt}
\end{figure}

\begin{figure}[t]
    \centering
    \includegraphics[width=0.8\linewidth]{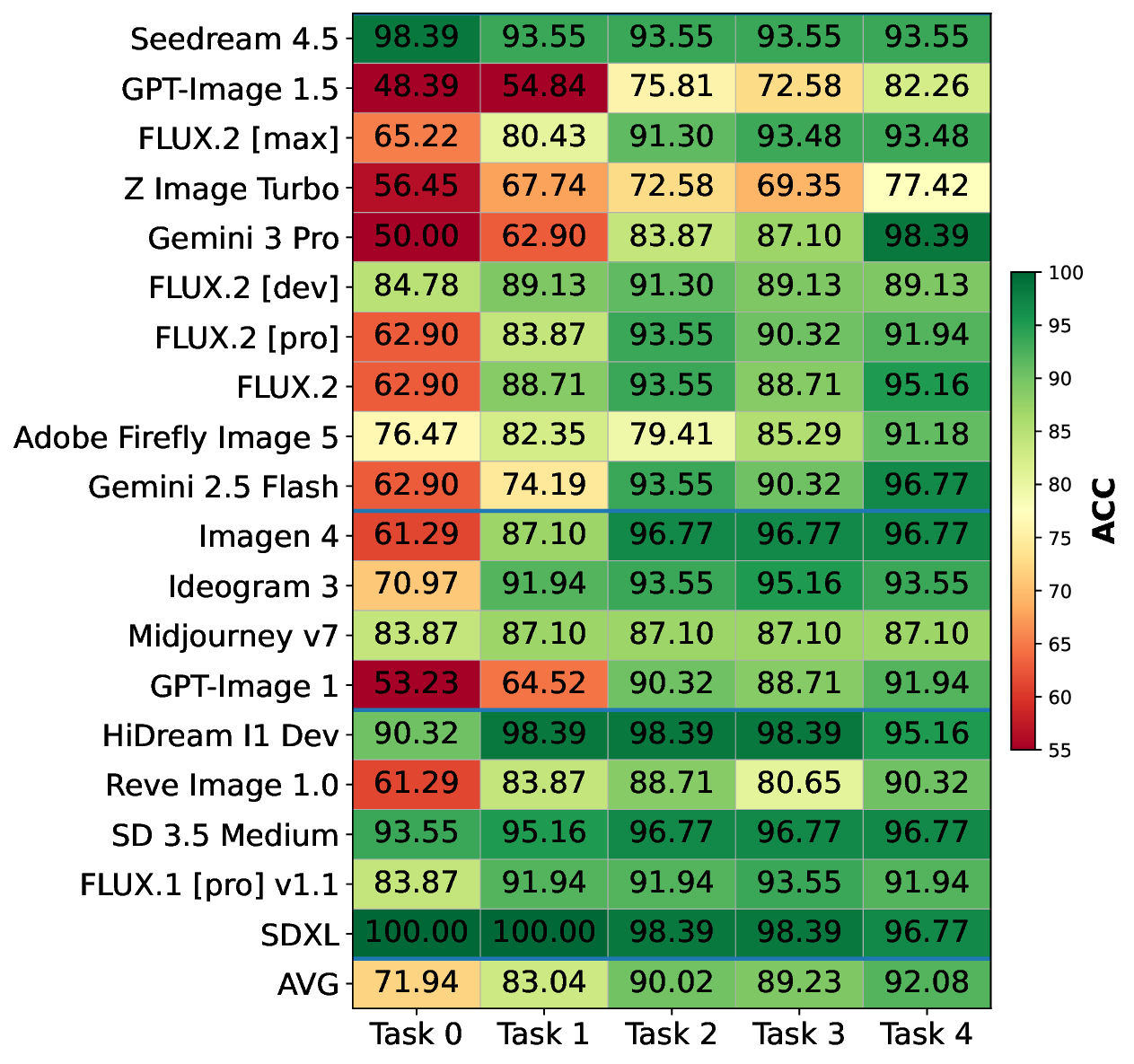}
    \vspace{-5pt}
    \caption{
    ACC on AIGenImages2026 across generators under continual learning. Results show consistent improvements across tasks, demonstrating effective adaptation while mitigating catastrophic forgetting.}
    \label{fig:continual_gen_acc}
    \vspace{-10pt}
\end{figure}

\begin{table*}[t]
\centering
\scriptsize
\setlength{\tabcolsep}{3.0pt}
\caption{Ablation study of the proposed framework components. Each entry reports AUC / ACC (\%).}
\vspace{-5pt}
\label{tab:rine_ablation}
\resizebox{0.9\textwidth}{!}{
\begin{tabular}{cc c| ccccccccccc}
\toprule
\multicolumn{2}{c}{\textbf{Training Data}} & \multirow{2}{*}{\textbf{Replay}}
& \multicolumn{5}{c}{\textbf{Recent Generators}}
& \multicolumn{4}{c}{\textbf{ITW Data}}
&  \\
\cmidrule(lr){1-2} \cmidrule(lr){4-8} \cmidrule(lr){9-12}

\textbf{WildFC} & \textbf{AI-Gen}
& 
& \textbf{AIGenImages2026}
& \textbf{MNW}
& \textbf{Echo-4o}
& \textbf{Nano-Consistent}
& \textbf{Synthbuster}
& \textbf{Chameleon}
& \textbf{DeepFake-Eval}
& \textbf{ITW-SM}
& \textbf{MediaEval}
& \textbf{AVG} \\
\midrule

\xmark & \xmark & \xmark
& 90.80 / 71.74
& 93.50 / 81.88
& 50.89 / 48.75
& 68.81 / 63.81
& 90.87 / 73.42
& 88.19 / \underline{79.60}
& 84.56 / 75.56
& 96.23 / 81.49
& 89.08 / 76.41
& 83.66 / 72.52 \\

\cmark & \xmark & \xmark
& 94.18 / 75.49
& 93.45 / 77.34
& 77.93 / 51.75
& \textbf{84.40} / \textbf{74.81}
& \underline{95.16} / \underline{85.62}
& 84.23 / 74.40
& 84.03 / 74.31
& 95.33 / \textbf{87.85}
& 89.47 / 80.13
& 88.69 / 75.74 \\

\cmark & \xmark & \textbf{5\%}
& 94.35 / 78.62
& 94.79 / 82.97
& 67.54 / 60.60
& \underline{79.53} / \underline{72.65}
& 93.64 / 80.30
& \textbf{90.93} / \textbf{81.04}
& 86.57 / 77.59
& \textbf{96.81} / \underline{85.14}
& 90.88 / \textbf{80.89}
& 88.34 / 77.76 \\

\xmark & \cmark & \textbf{5\%}
& \textbf{97.75} / 91.23
& \underline{95.85} / \underline{89.22}
& 89.25 / 75.05
& 72.84 / 66.99
& 95.08 / 75.55
& 88.54 / 75.90
& 86.40 / 77.28
& 95.61 / 78.98
& 90.21 / 75.55
& 90.17 / 78.42 \\

\cmark & \cmark & \xmark
& 95.74 / 87.30
& 90.97 / 82.44
& \textbf{98.73} / \textbf{90.10}
& 76.11 / 70.21
& \textbf{96.41} / \textbf{87.37}
& 78.86 / 66.24
& 84.57 / 76.45
& 90.60 / 80.52
& 85.21 / 77.03
& 88.58 / 79.74 \\

\cmark & \cmark & \textbf{3\%}
& \underline{97.57} / \underline{91.59}
& 95.18 / 87.12
& \underline{92.71} / \underline{82.10}
& 76.75 / 70.80
& 93.49 / 80.97
& 84.99 / 73.23
& 87.32 / \underline{78.43}
& 94.90 / 83.53
& 89.83 / \underline{80.22}
& 90.30 / \textbf{80.89} \\

\cmark & \cmark & \textbf{5\%}
& \textbf{97.75} / \textbf{92.13}
& 95.74 / 88.59
& 88.45 / 76.80
& 78.35 / 71.01
& 94.81 / 77.75
& 89.64 / 78.58
& \textbf{87.59} / \textbf{78.63}
& 95.74 / 82.50
& \underline{91.09} / 78.80
& \underline{91.02} / \underline{80.53} \\

\cmark & \cmark & \textbf{10\%}
& 97.42 / 91.06
& \textbf{95.89} / \textbf{89.36}
& 88.65 / 75.50
& 79.15 / 71.36
& 94.76 / 75.51
& \underline{89.89} / 75.91
& \underline{87.58} / 78.01
& \underline{95.80} / 78.65
& \textbf{91.13} / 76.61
& \textbf{91.14} / 79.11 \\

\bottomrule
\end{tabular}}
\vspace{-5pt}
\end{table*}

\begin{table*}[t]
\centering
\scriptsize
\setlength{\tabcolsep}{3.0pt}
\caption{Ablation study of the proposed framework across different training data portions. Each entry reports AUC / ACC (\%).}
\vspace{-5pt}
\label{tab:portion_comparison}
\resizebox{0.9\textwidth}{!}{
\begin{tabular}{lcccccccccc}
\toprule
\textbf{Size}
& \multicolumn{5}{c}{\textbf{Recent Generators}}
& \multicolumn{4}{c}{\textbf{ITW Data}}
& \textbf{} \\
\cmidrule(lr){2-6} \cmidrule(lr){7-10}

& \textbf{AIGenImages2026}
& \textbf{MNW}
& \textbf{Echo-4o}
& \textbf{Nano-Consistent}
& \textbf{Synthbuster}
& \textbf{Chameleon}
& \textbf{DeepFake-Eval}
& \textbf{ITW-SM}
& \textbf{MediaEval}
& \textbf{AVG} \\
\midrule

\textbf{0\%}
& 90.80 / 71.74
& 93.50 / 81.88
& 50.89 / 48.75
& 68.81 / 63.81
& 90.87 / 73.42
& 88.19 / \underline{79.60}
& 84.56 / 75.56
& \underline{96.23}/ 81.49
& 89.08 / 76.41
& 83.66 / 72.52 \\

\textbf{10\%}
& 94.07 / 77.82
& 94.54 / 82.92
& 62.44 / 58.05
& 72.49 / 65.42
& 90.53 / 75.91
& 89.62 / \textbf{81.70}
& 86.55 / \underline{78.06}
& \textbf{96.50} / \textbf{84.06}
& 89.97 / \underline{78.86}
& 86.30 / 75.87 \\

\textbf{30\%}
& 97.09 / \underline{90.88}
& \textbf{96.16} / \textbf{90.19}
& \underline{81.25} / \underline{72.95}
& \textbf{82.12} / 67.78
& \underline{94.29} / 74.77
& \textbf{91.12} / 73.28
& \underline{87.24} / 77.28
& 96.17 / 73.33
& \underline{90.60} / 71.29
& \underline{90.67} / 76.86 \\

\textbf{50\%}
& \underline{97.49} / 90.52
& \underline{95.98} / 86.64
& 77.55 / 72.90
& 77.30 / \underline{70.91}
& 92.72 / \textbf{79.19}
& 89.08 / 79.16
& 86.34 / 77.02
& 95.68 / \underline{83.39}
& 89.90 / \textbf{79.44}
& 89.12 / \underline{79.91} \\

\textbf{100\%}
& \textbf{97.75} / \textbf{92.13}
& 95.74 / \underline{88.59}
& \textbf{88.45} / \textbf{76.80}
& \underline{78.35} / \textbf{71.01}
& \textbf{94.81} / \underline{77.75}
& \underline{89.64} / 78.58
& \textbf{87.59} / \textbf{78.63}
& 95.74 / 82.50
& \textbf{91.09} / 78.80
& \textbf{91.02} / \textbf{80.53} \\

\bottomrule
\end{tabular}}
\vspace{-5pt}
\end{table*}

\subsection{Continual Learning Analysis}
We evaluate our framework by simulating a continual learning setting using the RINE model across four incremental tasks, defined at 3-month intervals throughout 2025. Starting from a pretrained model (Task 0), we progressively expand the training set with data accumulated up to March, June, September, and December 2025 for Tasks 1–4, respectively. Chronological ordering is based on WildFC image metadata and AIGenImages2026 release dates (see Table~\ref{tab:generators}). This setup constitutes a simulation, as samples from preview versions of generative models may appear earlier in in-the-wild data. A replay buffer is used during training. The results are presented in Fig.~\ref{fig:continual_auc} and Fig.~\ref{fig:continual_gen_acc}.

The pretrained RINE model (Task 0) performs poorly across several generators in AIGenImages2026 test set. Task 1, yields a clear improvement in average AUC across all benchmarks, with a gain of $+3.49\%$. On AIGenImages2026, the average accuracy increases by $+11.10\%$. Notably, substantial gains are observed for generators not directly included in training. For example, Gemini 3 Pro improves by $+12.90\%$ in terms of accuracy, FLUX.2 by $+25.81\%$, and Imagen 4 by $+25.81\%$, highlighting the strong generalization capability of the framework. 

A similar trend continues in Task 2, with average AUC rising another $+3.64\%$ over Task 1. We observe significant improvements to new generators, such as GPT-Image 1 with $+25.80\%$ in terms of accuracy and Gemini 2.5 Flash with $+19.36\%$. In parallel, in-the-wild performance improves compared to Task 0, including Chameleon with $+1.97\%$ in terms of AUC, Deepfake-Eval with $+3.30\%$, and MediaEval-ITW with $+1.49\%$. 

Task 3 maintains stable performance with a slight $-0.56\%$ decrease in average AUC. While AUC improves by $+0.33\%$ on ITW-SM and $+0.56\%$ on MediaEval-ITW, a minor degradation of $-0.16\%$ occurs on Deepfake-Eval, and a more noticeable drop of $-4.90\%$ appears in the Echo-4o dataset. Similarly, AIGenImages2026 shows a small overall accuracy decrease of $-0.79\%$. Despite this, improvements persist for several generators, including Gemini 3 Pro ($+13.23\%$ gain) and Adobe Firefly Image 5 ($+5.88\%$). In contrast, performance degrades for Reve Image 1.0 ($-8.06\%$) and Z Image Turbo ($-3.23\%$). 

Finally, Task 4 achieves the peak average AUC across all benchmarks, representing an improvement of $+7.36\%$ over Task 0. On AIGenImages2026, the proposed framework effectively adapts the RINE model, reaching a substantial $+20.14\%$ accuracy gain. Overall, these results confirm that the framework enables continual adaptation to evolving data distributions with only minor fluctuations, while mitigating catastrophic forgetting across both generators and in-the-wild data.

%\vspace{-0.3pt}
\subsection{Ablation Studies}
\textbf{Component contribution analysis}. We conduct an ablation study on WildFC, AIGenImages2026, and the replay mechanism (see Table~\ref{tab:rine_ablation}), using a RINE model pretrained on LDM and Twigma as the baseline. Training on WildFC without replay improves overall performance by $+5.03\%$ AUC and $+3.22\%$ ACC, but reduces Chameleon ACC by $-5.2\%$. When trained with replay, WildFC consistently improves in-the-wild performance, including $+1.44\%$ on Chameleon, $+2.03\%$ on Deepfake-Eval, $+3.65\%$ on ITW-SM, and $+4.48\%$ on MediaEval-ITW, while also generalizing to recent generators, with gains of $+6.88\%$ on AIGenImages2026, $+1.09\%$ on MNW, $+11.85\%$ on Echo-4o, and $+8.84\%$ on Nano-Consistent in ACC, despite no access to generator-specific training data. In contrast, training on AIGenImages2026 with replay substantially improves generator performance, yielding $+19.49\%$ on AIGenImages2026 and $+7.34\%$ on MNW, but degrades in-the-wild generalization, with drops of $-3.70\%$ on Chameleon, $-2.51\%$ on ITW-SM, and $-0.86\%$ on MediaEval-ITW, indicating a specialization bias. Combining both datasets without replay further strengthens generator performance, including $+15.56\%$ on Echo-4o and $+13.95\%$ on Synthbuster, but significantly harms in-the-wild robustness, with decreases of $-13.36\%$ on Chameleon and $-4.07\%$ on ITW-SM, highlighting distribution interference between real and synthetic data. Importantly, the replay buffer is crucial for mitigating catastrophic forgetting. A $3\%$ buffer improves performance across all in-the-wild datasets while also increasing performance on AIGenImages2026 by $+4.29\%$ and MNW by $+4.68\%$. The $5\%$ buffer yields the best trade-off, recovering in-the-wild performance by $+12.34\%$ on Chameleon and $+2.18\%$ on Deepfake-Eval, and achieving improvements of $+7.36\%$ AUC and $+8.01\%$ ACC over the baseline. Increasing the buffer to $10\%$ provides no meaningful gain, with changes of $-1.42\%$ ACC and $+0.12\%$ AUC, so we adopt $5\%$ for efficiency. Overall, in-the-wild data improves robustness, synthetic data enhances specialization, and replay enables their effective integration.

\textbf{Effect of training data size}. We analyze the effect of training data size by progressively increasing the proportion of AIGenImages2026 and WildFC used to train the RINE model, as shown in Table~\ref{tab:portion_comparison}. Even 10\% of the data yields substantial improvements, with gains of $+11.55\%$ on Echo-4o, $+3.68\%$ on Nano-Consistent, and $+3.35\%$ on average AUC. At 30\%, the model achieves strong performance on generator datasets, with $+6.71\%$ on MNW and $+24.2\%$ on Echo-4o, while showing small improvements on in-the-wild data with $+2.92\%$ on Chameleon and $+2.68\%$ on Deepfake-Eval in AUC. However, this comes with instability in in-the-wild ACC, indicating suboptimal calibration rather than reduced discriminative power. Increasing the data to 50\% improves stability and raises average accuracy by $+3.05\%$ over the 30\% setting. Using 100\% of the data yields the best performance. Overall, these results suggest that the model adapts to new distributions even with limited data, while larger training sets consistently improve performance.

\section{Limitations and Future Work}

While our approach achieves strong performance several limitations remain. The fact-check retrieval pipeline relies on weak supervision and web-sourced data, which inevitably introduces noise. Incorrect labeling may affect dataset quality and the generalization ability of the model. Future work should focus on refined curation and scaling data collection efficiency. Furthermore, while we treat generator-driven and in-the-wild data as separate sources, developing principled strategies to integrate them could better approximate real-world distributions. Extensions could explore automated image discovery for emerging models, more advanced continual learning strategies, and multimodal content.

\section{Conclusion}
\label{sec:conclusion}
In this paper, we introduce a continual adaptive framework for AID, addressing the challenge of distribution shift in a rapidly evolving generative content. Unlike prior approaches that rely on static datasets or generator-specific adaptation, our method models AID as a dynamic problem and enables continuous adaptation. To support this, we propose a fact-check retrieval pipeline that enables automated, weakly supervised dataset construction from real-world sources. We further introduce an evolving dataset designed to reflect both emerging generative models and real-world data distributions. Extensive experiments across multiple benchmarks and two different detection architectures demonstrate the general applicability of our framework, consistently achieving state-of-the-art performance. Our results show that combining in-the-wild and generator-driven data, together with a continual data collection and learning strategy, significantly improves generalization to both recent generative models and real-world scenarios.
\clearpage

\section*{Acknowledgments}
This work received funding by the Horizon Europe projects AI-CODE (GA no. 101135437) and ELIAS (GA no. 101120237).

\bibliographystyle{unsrt}
\bibliography{sample-base}

@String{Computer = "{IEEE} Computer" }

@String{Springer = "Springer-Verlag" }

@article{moreno2012unifying,
  title={A unifying view on dataset shift in classification},
  author={Moreno-Torres, Jose G and Raeder, Troy and Alaiz-Rodr{\'\i}guez, Roc{\'\i}o and Chawla, Nitesh V and Herrera, Francisco},
  journal={Pattern recognition},
  volume={45},
  number={1},
  pages={521--530},
  year={2012},
  publisher={Elsevier}
}

@inproceedings{yan2025sanity,
  title={A Sanity Check for AI-generated Image Detection},
  author={Yan, Shilin and Li, Ouxiang and Cai, Jiayin and Hao, Yanbin and Jiang, Xiaolong and Hu, Yao and Xie, Weidi},
  booktitle={Proceedings of the International Conference on Learning Representations (ICLR)},
  year={2025}
}

@inproceedings{wang2020cnn,
  title={CNN-generated images are surprisingly easy to spot... for now},
  author={Wang, Sheng-Yu and Wang, Oliver and Zhang, Richard and Owens, Andrew and Efros, Alexei A},
  booktitle={Proceedings of the IEEE/CVF conference on computer vision and pattern recognition},
  pages={8695--8704},
  year={2020}
}

@inproceedings{tan2024rethinking,
  title={Rethinking the up-sampling operations in cnn-based generative network for generalizable deepfake detection},
  author={Tan, Chuangchuang and Zhao, Yao and Wei, Shikui and Gu, Guanghua and Liu, Ping and Wei, Yunchao},
  booktitle={Proceedings of the IEEE/CVF conference on computer vision and pattern recognition},
  pages={28130--28139},
  year={2024}
}

@inproceedings{li2025improving,
  title={Improving synthetic image detection towards generalization: An image transformation perspective},
  author={Li, Ouxiang and Cai, Jiayin and Hao, Yanbin and Jiang, Xiaolong and Hu, Yao and Feng, Fuli},
  booktitle={Proceedings of the 31st ACM SIGKDD Conference on Knowledge Discovery and Data Mining V. 1},
  pages={2405--2414},
  year={2025}
}

@article{ye2025echo,
  title={Echo-4o: Harnessing the power of gpt-4o synthetic images for improved image generation},
  author={Ye, Junyan and Jiang, Dongzhi and Wang, Zihao and Zhu, Leqi and Hu, Zhenghao and Huang, Zilong and He, Jun and Yan, Zhiyuan and Yu, Jinghua and Li, Hongsheng and He, Conghui and Li, Weijia},
  journal={arXiv preprint arXiv:2508.09987},
  year={2025}
}

@article{cavia2024real,
  title={Real-time deepfake detection in the real-world},
  author={Cavia, Bar and Horwitz, Eliahu and Reiss, Tal and Hoshen, Yedid},
  journal={arXiv preprint arXiv:2406.09398},
  year={2024}
}

@inproceedings{koutlis2024leveraging,
  title={Leveraging representations from intermediate encoder-blocks for synthetic image detection},
  author={Koutlis, Christos and Papadopoulos, Symeon},
  booktitle={European Conference on computer vision},
  pages={394--411},
  year={2024},
  organization={Springer}
}

@inproceedings{karageorgiou2025any,
  title={Any-resolution ai-generated image detection by spectral learning},
  author={Karageorgiou, Dimitrios and Papadopoulos, Symeon and Kompatsiaris, Ioannis and Gavves, Efstratios},
  booktitle={Proceedings of the Computer Vision and Pattern Recognition Conference},
  pages={18706--18717},
  year={2025}
}

@article{konstantinidou2025navigating,
  title={Navigating the Challenges of AI-Generated Image Detection in the Wild: What Truly Matters?},
  author={Konstantinidou, Despina and Karageorgiou, Dimitrios and Koutlis, Christos and Papadopoulou, Olga and Schinas, Emmanouil and Papadopoulos, Symeon},
  journal={arXiv preprint arXiv:2507.10236},
  year={2025}
}

@inproceedings{guillaro2025bias,
  title={A bias-free training paradigm for more general ai-generated image detection},
  author={Guillaro, Fabrizio and Zingarini, Giada and Usman, Ben and Sud, Avneesh and Cozzolino, Davide and Verdoliva, Luisa},
  booktitle={Proceedings of the Computer Vision and Pattern Recognition Conference},
  pages={18685--18694},
  year={2025}
}

@inproceedings{cozzolino2024raising,
  title={Raising the bar of ai-generated image detection with clip},
  author={Cozzolino, Davide and Poggi, Giovanni and Corvi, Riccardo and Nie{\ss}ner, Matthias and Verdoliva, Luisa},
  booktitle={Proceedings of the IEEE/CVF Conference on Computer Vision and Pattern Recognition},
  pages={4356--4366},
  year={2024}
}

@article{chen2025opengpt,
  title={Opengpt-4o-image: A comprehensive dataset for advanced image generation and editing},
  author={Chen, Zhihong and Bai, Xuehai and Shi, Yang and Fu, Chaoyou and Zhang, Huanyu and Wang, Haotian and Sun, Xiaoyan and Zhang, Zhang and Wang, Liang and Zhang, Yuanxing and others},
  journal={arXiv preprint arXiv:2509.24900},
  year={2025}
}

@inproceedings{li2025bridging,
  title={Bridging the Gap Between Ideal and Real-world Evaluation: Benchmarking AI-Generated Image Detection in Challenging Scenarios},
  author={Li, Chunxiao and Wang, Xiaoxiao and Li, Meiling and Miao, Boming and Sun, Peng and Zhang, Yunjian and Ji, Xiangyang and Zhu, Yao},
  booktitle={Proceedings of the IEEE/CVF International Conference on Computer Vision},
  pages={20379--20389},
  year={2025}
}

@inproceedings{pellegrini2025ai,
  title={AI-GenBench: A New Ongoing Benchmark for AI-Generated Image Detection},
  author={Pellegrini, Lorenzo and Cozzolino, Davide and Pandolfini, Serafino and Maltoni, Davide and Ferrara, Matteo and Verdoliva, Luisa and Prati, Marco and Ramilli, Marco},
  booktitle={2025 International Joint Conference on Neural Networks (IJCNN)},
  pages={1--9},
  year={2025},
  organization={IEEE}
}

@article{li2025artificial,
  title={Is artificial intelligence generated image detection a solved problem?},
  author={Li, Ziqiang and Yan, Jiazhen and He, Ziwen and Zeng, Kai and Jiang, Weiwei and Xiong, Lizhi and Fu, Zhangjie},
  journal={arXiv preprint arXiv:2505.12335},
  year={2025}
}

@inproceedings{yu2025realhd,
  title={RealHD: A High-Quality Dataset for Robust Detection of State-of-the-Art AI-Generated Images},
  author={Yu, Hanzhe and Ye, Yun and Rong, Jintao and Xuan, Qi and Ma, Chen},
  booktitle={Proceedings of the 33rd ACM International Conference on Multimedia},
  pages={11394--11403},
  year={2025}
}

@inproceedings{doherty2024clofai,
  title={CLOFAI: A Dataset of Real And Fake Image Classification Tasks for Continual Learning},
  author={Doherty, William and Lee, Anton and Gomes, Heitor Murilo},
  booktitle={International Conference on Neural Information Processing},
  pages={348--362},
  year={2024},
  organization={Springer}
}

@article{chen2023twigma,
  title={Twigma: A dataset of ai-generated images with metadata from twitter},
  author={Chen, Yiqun and Zou, James Y},
  journal={Advances in Neural Information Processing Systems},
  volume={36},
  pages={37748--37760},
  year={2023}
}

@inproceedings{corvi2023detection,
  title={On the detection of synthetic images generated by diffusion models},
  author={Corvi, Riccardo and Cozzolino, Davide and Zingarini, Giada and Poggi, Giovanni and Nagano, Koki and Verdoliva, Luisa},
  booktitle={ICASSP 2023-2023 IEEE International Conference on Acoustics, Speech and Signal Processing (ICASSP)},
  pages={1--5},
  year={2023},
  organization={IEEE}
}

@article{ye2025realgen,
  title={Realgen: Photorealistic text-to-image generation via detector-guided rewards},
  author={Ye, Junyan and Zhu, Leiqi and Guo, Yuncheng and Jiang, Dongzhi and Huang, Zilong and Zhang, Yifan and Yan, Zhiyuan and Fu, Haohuan and He, Conghui and Li, Weijia},
  journal={arXiv preprint arXiv:2512.00473},
  year={2025}
}

@inproceedings{papadopoulou2025synthetic,
  title={Synthetic images at mediaeval 2025: Advancing detection of generative ai in real-world online images},
  author={Papadopoulou, Olga and Schinas, Manos and Corvi, Riccardo and Karageorgiou, Dimitrios and Koutlis, Christos and Guillaro, Fabrizio and Gavves, Efstratios and Mareen, Hannes and Verdoliva, Luisa and Papadopoulos, Symeon},
  booktitle={Proceedings of the MediaEval 2025 Workshop, Dublin, Ireland and Online},
  pages={25--26},
  year={2025}
}

@article{wang2022s,
  title={S-prompts learning with pre-trained transformers: An occam’s razor for domain incremental learning},
  author={Wang, Yabin and Huang, Zhiwu and Hong, Xiaopeng},
  journal={Advances in Neural Information Processing Systems},
  volume={35},
  pages={5682--5695},
  year={2022}
}

@inproceedings{epstein2023online,
  title={Online detection of ai-generated images},
  author={Epstein, David C and Jain, Ishan and Wang, Oliver and Zhang, Richard},
  booktitle={Proceedings of the IEEE/CVF international conference on computer vision},
  pages={382--392},
  year={2023}
}

@inproceedings{azizpour2024e3,
  title={E3: Ensemble of expert embedders for adapting synthetic image detectors to new generators using limited data},
  author={Azizpour, Aref and Nguyen, Tai D and Shrestha, Manil and Xu, Kaidi and Kim, Edward and Stamm, Matthew C},
  booktitle={Proceedings of the IEEE/CVF Conference on Computer Vision and Pattern Recognition},
  pages={4334--4344},
  year={2024}
}

@article{lu2025liteupdate,
  title={LiteUpdate: A Lightweight Framework for Updating AI-Generated Image Detectors},
  author={Lu, Jiajie and Fu, Zhenkan and Zhao, Na and Xing, Long and Chen, Kejiang and Zhang, Weiming and Yu, Nenghai},
  journal={arXiv preprint arXiv:2511.07192},
  year={2025}
}

@article{pellegrini2025generalized,
  title={Generalized Design Choices for Deepfake Detectors},
  author={Pellegrini, Lorenzo and Pandolfini, Serafino and Maltoni, Davide and Ferrara, Matteo and Prati, Marco and Ramilli, Marco},
  journal={arXiv preprint arXiv:2511.21507},
  year={2025}
}

@inproceedings{liu2021visual,
  title={Visual news: Benchmark and challenges in news image captioning},
  author={Liu, Fuxiao and Wang, Yinghan and Wang, Tianlu and Ordonez, Vicente},
  booktitle={Proceedings of the 2021 conference on empirical methods in natural language processing},
  pages={6761--6771},
  year={2021}
}

@article{chandra2025deepfake,
  title={Deepfake-eval-2024: A multi-modal in-the-wild benchmark of deepfakes circulated in 2024},
  author={Chandra, Nuria Alina and Murtfeldt, Ryan and Qiu, Lin and Karmakar, Arnab and Lee, Hannah and Tanumihardja, Emmanuel and Farhat, Kevin and Caffee, Ben and Paik, Sejin and Lee, Changyeon and others},
  journal={arXiv preprint arXiv:2503.02857},
  year={2025}
}

@misc{MNW2025,
      title={Introducing the MNW benchmark for AI forensics}, 
      author={Thomas Roca and Marco Postiglione and Chongyang Gao and Isabel Gortner and Zuzanna Wojciak and Pengce Wang and Mahsa Alimardani and Shirin Anlen and Kevin White and Juan Lavista Ferres and Sarit Kraus and Sam Gregory and V. S. Subrahmanian},
      year={2025},
      eprint={......},
      archivePrefix={arXiv},
}

@article{bammey2023synthbuster,
  title={Synthbuster: Towards detection of diffusion model generated images},
  author={Bammey, Quentin},
  journal={IEEE Open Journal of Signal Processing},
  volume={5},
  pages={1--9},
  year={2023},
  publisher={IEEE}
}

@inproceedings{schinas2024sidbench,
  title={SIDBench: A Python framework for reliably assessing synthetic image detection methods},
  author={Schinas, Manos and Papadopoulos, Symeon},
  booktitle={Proceedings of the 3rd ACM International Workshop on Multimedia AI against Disinformation},
  pages={55--64},
  year={2024}
}

@article{paszke2019pytorch,
  title={Pytorch: An imperative style, high-performance deep learning library},
  author={Paszke, Adam and Gross, Sam and Massa, Francisco and Lerer, Adam and Bradbury, James and Chanan, Gregory and Killeen, Trevor and Lin, Zeming and Gimelshein, Natalia and Antiga, Luca and others},
  journal={Advances in neural information processing systems},
  volume={32},
  year={2019}
}

@inproceedings{karageogiou2024evolution,
  title={Evolution of detection performance throughout the online lifespan of synthetic images},
  author={Karageogiou, Dimitrios and Bammey, Quentin and Porcellini, Valentin and Goupil, Bertrand and Teyssou, Denis and Papadopoulos, Symeon},
  booktitle={European Conference on Computer Vision},
  pages={400--417},
  year={2024},
  organization={Springer}
}

@article{sengar2025generative,
  title={Generative artificial intelligence: a systematic review and applications},
  author={Sengar, Sandeep Singh and Hasan, Affan Bin and Kumar, Sanjay and Carroll, Fiona},
  journal={Multimedia Tools and Applications},
  volume={84},
  number={21},
  pages={23661--23700},
  year={2025},
  publisher={Springer}
}

@article{yazdani2025generative,
  title={Generative AI in depth: A survey of recent advances, model variants, and real-world applications},
  author={Yazdani, Shamim and Singh, Akansha and Saxena, Nripsuta and Wang, Zichong and Palikhe, Avash and Pan, Deng and Pal, Umapada and Yang, Jie and Zhang, Wenbin},
  journal={Journal of Big Data},
  volume={12},
  number={1},
  pages={230},
  year={2025},
  publisher={Springer}
}

@inproceedings{radford2021learning,
  title={Learning transferable visual models from natural language supervision},
  author={Radford, Alec and Kim, Jong Wook and Hallacy, Chris and Ramesh, Aditya and Goh, Gabriel and Agarwal, Sandhini and Sastry, Girish and Askell, Amanda and Mishkin, Pamela and Clark, Jack and others},
  booktitle={International conference on machine learning},
  pages={8748--8763},
  year={2021},
  organization={PmLR}
}

@misc{google_factcheck_api,
  author = {{Google}},
  title = {Google Fact Check Tools API},
  year = {2026},
  url = {https://developers.google.com/fact-check/tools/api},
  note = {Accessed: 4 Apr. 2026}
}

@misc{dbkf,
  author = {{Ontotext}},
  title = {The Database of Known Fakes (DBKF)},
  year = {2026},
  url = {https://dbkf.ontotext.com},
  note = {Accessed: 4 Apr. 2026}
}

@misc{crawl4ai,
  author = {{UncleCode}},
  title = {Crawl4AI: An Open-Source LLM-Friendly Web Crawler and Scraper},
  year = {2024},
  url = {https://github.com/unclecode/crawl4ai},
  note = {Accessed: 4 Apr. 2026}
}

@misc{qwen3technicalreport,
      title={Qwen3 Technical Report}, 
      author={Qwen Team},
      year={2025},
      eprint={2505.09388},
      archivePrefix={arXiv},
      primaryClass={cs.CL},
      url={https://arxiv.org/abs/2505.09388}, 
}

@misc{qwen25_vl,
  author = {{Qwen Team}},
  title = {Qwen2.5-VL},
  year = {2025},
  url = {https://qwenlm.github.io/blog/qwen2.5-vl/},
  note = {Accessed: 4 Apr. 2026}
}

@inproceedings{liu2024grounding,
  title={Grounding dino: Marrying dino with grounded pre-training for open-set object detection},
  author={Liu, Shilong and Zeng, Zhaoyang and Ren, Tianhe and Li, Feng and Zhang, Hao and Yang, Jie and Jiang, Qing and Li, Chunyuan and Yang, Jianwei and Su, Hang and others},
  booktitle={European conference on computer vision},
  pages={38--55},
  year={2024},
  organization={Springer}
}

@misc{gpt52,
  author = {{OpenAI}},
  title = {Introducing GPT-5.2},
  year = {2026},
  url = {https://openai.com/index/introducing-gpt-5-2/},
  note = {Accessed: 4 Apr. 2026}
}

@misc{falai,
  author = {{FAL.ai}},
  title = {FAL.ai API for Generative Image Models},
  year = {2026},
  url = {https://fal.ai},
  note = {Accessed: 4 Apr. 2026}
}

@article{oquab2023dinov2,
  title={Dinov2: Learning robust visual features without supervision},
  author={Oquab, Maxime and Darcet, Timoth{\'e}e and Moutakanni, Th{\'e}o and Vo, Huy and Szafraniec, Marc and Khalidov, Vasil and Fernandez, Pierre and Haziza, Daniel and Massa, Francisco and El-Nouby, Alaaeldin and others},
  journal={arXiv preprint arXiv:2304.07193},
  year={2023}
}

@misc{gallerydl,
  author = {{Faehrmann, Mike}},
  title = {gallery-dl},
  year = {2026},
  url = {https://github.com/mikf/gallery-dl},
  note = {Accessed: 5 Apr. 2026}
}

@misc{newsapi,
  author = {{NewsAPI}},
  title = {NewsAPI},
  year = {2026},
  url = {https://newsapi.org/},
  note = {Accessed: 5 Apr. 2026}
}

\end{document}